\pdfoutput=1
\documentclass[11pt,a4paper]{article}
\usepackage[hyperref]{acl2020}
\usepackage{times}
\usepackage{latexsym}

\usepackage{microtype}

\aclfinalcopy 

\usepackage{graphicx}
\usepackage{verbatim}
\usepackage{siunitx}
\usepackage[shortlabels]{enumitem}
\usepackage{tabularx}
\usepackage{booktabs}
\usepackage{caption}
\usepackage{colortbl}
\usepackage{subcaption}
\usepackage{multirow}

\title{STARC: Structured Annotations for Reading Comprehension}

\author{Yevgeni Berzak \\
  MIT BCS\\
  \texttt{berzak@mit.edu} \\\And
  Jonathan Malmaud \\
  MIT BCS\\
  \texttt{malmaud@mit.edu} \\\And
  Roger Levy \\
  MIT BCS\\
  \texttt{rplevy@mit.edu} \\}

\date{}

\begin{document}
\maketitle
\begin{abstract}
       We present STARC (Structured Annotations for Reading Comprehension), a new annotation framework for assessing reading comprehension with multiple choice questions. Our framework introduces a principled structure for the answer choices and ties them to textual span annotations. The framework is implemented in OneStopQA, a new high-quality dataset for evaluation and analysis of reading comprehension in English. We use this dataset to demonstrate that STARC can be leveraged for a key new application for the development of SAT-like reading comprehension materials: automatic annotation quality probing via span ablation experiments. We further show that it enables in-depth analyses and comparisons between machine and human reading comprehension behavior, including error distributions and guessing ability. Our experiments also reveal that the standard multiple choice dataset in NLP, RACE \cite{lai2017race}, is limited in its ability to measure reading comprehension. 47\% of its questions can be guessed by machines without accessing the passage, and 18\% are unanimously judged by humans as not having a unique correct answer. OneStopQA provides an alternative test set for reading comprehension which alleviates these shortcomings and has a substantially higher human ceiling performance.\footnote{OneStopQA dataset, STARC guidelines and human experiments data are available at \url{https://github.com/berzak/onestop-qa}}
\end{abstract}

\section{Introduction}

Assessment of reading comprehension is of paramount importance in education and science and is a key component of high-stakes evaluations such as the SAT examinations. Reading comprehension tasks are also central to NLP, where extensive efforts are invested in developing systems that try to match human-level performance. Despite the proliferation of NLP work on reading comprehension and the increasing number of large-scale reading comprehension datasets, key quality assurance issues such as question guessability, 
unwanted dataset biases, and the considerable success of simple pattern matching and slot filling heuristics remain open challenges for ensuring that evaluation benchmarks capture genuine reading comprehension. Further, existing annotation frameworks have very limited support for reading behavior analyses which go beyond simple accuracy statistics.

In this work, we introduce \textbf{STARC}, a new annotation framework for multiple choice reading comprehension, which addresses these shortcomings. Our framework aims to ensure high annotation quality and supports detailed probing and comparisons of human and machine reading comprehension behavior. The following are the primary novel characteristics of our annotation scheme.

\textbf{Structured Answer Choices} As opposed to existing multiple choice reading comprehension datasets, our framework has a principled and consistent answer structure. 
Specifically, every question has four possible answers. The first answer is the correct answer. Importantly, the correct answer typically does not appear verbatim in the passage. The second answer represents a misunderstanding of the critical information for answering the question correctly. The third answer refers to  information in the passage that is not relevant for the question. The fourth distractor has no support in the passage. This structure reflects four fundamental types of responses, ordered by miscomprehension severity.

\textbf{Auxiliary Span Annotations} 
To further enhance the versatility of the annotation scheme, the framework provides span annotations for the different answer choices. This approach creates a systematic correspondence between answers and their textual support. Specifically, the correct answer relies on a \textit{critical span} which contains the essential information for answering the question. In contrast to span identification datasets such as SQUAD \cite{rajpurkar2016squad} and Natural Questions \cite{naturalquestions2019}, we do not consider the span as the correct answer, but rather as a text region that contains the critical information required for answering the question correctly. The second answer represents a misunderstanding of that same span. Finally, the information referred to in the third answer is marked in a \textit{distractor span}. In this paper we demonstrate that the combination of a consistent answer structure with span annotations opens the door for new approaches to automatic verification of annotations and enables new types of analyses for reading comprehension.

We further introduce \textbf{OneStopQA}, a new dataset for multiple choice reading comprehension which implements our annotation framework. OneStopQA is a carefully constructed high-quality dataset intended primarily for testing and analyses, thereby complementing the existing larger multiple choice dataset RACE \cite{lai2017race}, which also has a 4-answer format and is commonly used for training. OneStopQA is designed to be challenging for both machine and human readers. The dataset comprises 30 articles from the Guardian in three parallel text difficulty versions and contains 1,458 paragraph-question pairs with multiple choice questions, along with manual span markings for both correct and incorrect answers. Despite its shorter passages and more constrained annotation scheme, baselines perform worse on OneStopQA than on RACE and the performance of a state-of-the-art model is comparable on both datasets.



We use OneStopQA to introduce an ablation-based framework for \emph{automatic verification of multiple choice reading comprehension materials} and to measure the extent to which the dataset can be solved without performing reading comprehension. Our framework is inspired by prior work on tasks such as image captioning and Visual Question Answering (VQA), where models were shown to perform well despite limited reliance on the images or the questions \cite{jabri2016,agrawal2016vqa,2017vqa,chao2018}. We utilize this framework to demonstrate the validity of OneStopQA annotations and their robustness to heuristics. 

Our analyses further reveal quality control issues in RACE. Machine readers are able to guess the correct answers to 47.1\% of the questions in RACE without being exposed to the passage, as opposed to 37.2\% for OneStopQA. When presented to humans via crowdsourcing, 18.3\% of the questions in RACE are unanimously judged by three annotators as not having a single correct answer, compared to only 3.4\% for OneStopQA. Using this human data, we establish an approximate ceiling above which model performance improvements are not likely to be meaningful: 88.8\% on RACE and 97.9\% on OneStopQA. We further verify this ceiling approximation with an in-lab human reading comprehension experiment in which we obtain a superior empirical human ceiling of 95.3\% for OneStopQA as compared to 84.7\% for RACE. These results are consequential in that state-of-the-art models are already around ceiling performance on RACE, while substantial room for improvement is still available for OneStopQA.

Finally, we showcase how the structure of OneStopQA annotations can be used for detailed comparisons between human and machine readers. Specifically, we demonstrate that human subjects and a state-of-the-art machine reading comprehension model have similar distributions of erroneous answers, suggesting a deeper link between human and machine readers than previously reported. On the other hand, humans and machines are fundamentally different in their guessing behavior. 

To summarize, the primary contributions of this work are the following:
\begin{itemize}
    \item We present STARC, an annotation framework for reading comprehension which combines structured answers with span annotations for both correct answers and distractors.
    \item We annotate and release OneStopQA, a dataset which adheres to this framework.
    \item We introduce a new methodology which leverages our annotations for automated data quality probing via ablation experiments.
    \item We showcase the value of the annotation framework for detailed analyses of human and machine reading comprehension behavior.
    \item Our experiments reveal that RACE is highly guessable and has a relatively low human ceiling due to low item quality in a large portion of the questions. OneStopQA does not have these drawbacks and can serve as an alternative out-of-domain challenge dataset for evaluations, compatible with training on RACE.
\end{itemize}
The combination of the novel annotation framework and the presented experiments suggests that the proposed annotation framework and our dataset can improve both the depth and the breadth of reading comprehension evaluations. 

\section{STARC Annotation Scheme}
\label{sec:guidelines-overview}

STARC is a new annotation framework accompanied by a protocol for increasing annotation quality and reducing annotation biases which can be exploited by either humans or machines for solving reading comprehension datasets without performing the intended task. The annotation scheme aims for the questions to be on a high difficulty level. Importantly, STARC tries to minimize the possibility of answering questions correctly using simple string-matching strategies, as well as guessing the correct answer without reading the passage. To focus on testing language comprehension, as opposed to other types of skills and knowledge, it aims to  avoid questions that rely on numerical reasoning and substantial external world knowledge. It also refrains from questions that require the reader to speculate (for example, given some information on person X, ask about their likely position issue Y when this position is not stated in the text). 

Reading comprehension questions have four answers, structured in the following manner.  \\
\textbf{A} is the correct answer. Answering a question correctly requires comprehending information from a text span in the passage called the \emph{critical span}. Importantly, with exceptions when necessary, the correct answer should not appear in the critical span in verbatim form. \\
\textbf{B} is an incorrect answer which represents a plausible misunderstanding of the critical span. \\ 
\textbf{C} is an incorrect answer which refers to an additional span in the passage, called the \emph{distractor span}. This answer can be anchored in the distractor span in various ways. For example, it may borrow keywords, or contain a correct fact that is stated in the distractor span but is not the correct answer to the question. \\ 
\textbf{D} is an incorrect answer which is plausible a-priori, but has no support in the passage. Note that to be plausible, D often appeals to the reader's general world knowledge. \\
Neither the critical span nor the distractor span have to adhere to sentence boundaries, and both can be non-continuous. 

This structure introduces well-defined and consistent relations between the answers and the passage. Further, the answers are ordered by degree of comprehension, whereby A represents correct comprehension, B reflects the ability to identify the crucial information for answering the question but failure to comprehend it, C reflects some degree of attention to the passage's content, and D provides no evidence for text comprehension. The utilization of B-type answers in particular enables probing comprehension at a deep level. The overall answer structure can support new types of error analyses beyond the correct/incorrect distinction by examining specific types of miscomprehension and their relation to the text. 

In order to reduce the effectiveness of answer elimination strategies, we developed additional guidelines on the joint form and content of the answers. These include a quality ranking of answer patterns, where the most preferred structures are those in which all answers have either similar phrasings or distinct phrasings. For all other patterns (e.g. three similarly worded answers and an outstanding answer), the answer types for the pattern should be distributed equally across questions. The guidelines also list dispreferred content relations between answers, such as B being the opposite of A. Finally, the guidelines specify that the answers across, and whenever possible within questions should be of comparable length. 

\section{OneStopQA Dataset}
\label{sec:dataset-overview}

\begin{table*}[ht!]
\small
\begin{center}
\begin{tabular}{l|ll|lll}
\cline{2-6}
 & \multicolumn{2}{c}{\bf RACE}& \multicolumn{3}{|c}{\bf OneStopQA} \\ \cline{2-6}
  & Middle & High & Ele & Int & Adv \\ \hline
Passages &6,409 / 368 / 362 & 18,728 / 1,021 / 1,045 & 162 & 162 & 162 \\
Questions&25,421 / 1,436 / 1,436 & 62,445 / 3,451 / 3,498 & 486 & 486 & 486 \\
 Words per passage & 232.12 & 354.08 & 112.32 & 126.97 & 138.6 \\
 Sentences per passage & 16.6 & 17.99 & 5.42 & 5.4 & 5.36 \\
 Words per sentence & 13.99 & 19.69 & 20.72 & 23.53 & 25.84 \\ \hline
 Flesh Kincaid & 3.24 & 7.06 & 7.32 & 8.9 & 10.1 \\
 SMOG & 7.58& 10.14 & 10.29 & 11.4 & 12.21 \\
\hline
\end{tabular}
\end{center}
\caption{\label{tab:onestop-stats-table} RACE and OneStopQA corpus statistics. The term ``passage'' refers to a single paragraph in OneStopQA and a single article in RACE. Values for the number of RACE passages and questions are formatted as Train / Dev / Test, while the remaining RACE values are calculated across the entire dataset. The readability measures Flesh Kincaid \cite{kincaid1975} and SMOG \cite{smog1969} are heuristic estimates of the number of education years required to fully comprehend the text.}
\end{table*}

\begin{table*}[ht!]
\small
\begin{tabularx}{\linewidth}{ l|X } 
\hline
    \bf Advanced&  A major international disagreement with wide-ranging implications for global drugs policy has erupted over the right of Bolivia's indigenous Indian tribes to chew coca leaves, the principal ingredient in cocaine. \textbf{\textcolor{red}{Bolivia has obtained a special exemption from the 1961 Single Convention on Narcotic Drugs, the framework that governs international drugs policy, allowing its indigenous people to chew the leaves.}} \textit{\textcolor{blue}{Bolivia had argued that the convention was in opposition to its new constitution,}} adopted in 2009, which obliges it to ``protect native and ancestral coca as cultural patrimony'' and maintains that coca ``in its natural state ... is not a dangerous narcotic.''\\ \hline
    \bf Elementary & A big international disagreement has started over the right of Bolivia's indigenous Indian tribes to chew coca leaves, the main ingredient in cocaine. This could have a significant effect on global drugs policy. \textbf{\textcolor{red}{Bolivia has received a special exemption from the 1961 Convention on Drugs, the agreement that controls international drugs policy. The exemption allows Bolivia's indigenous people to chew the leaves.}} \textit{\textcolor{blue}{Bolivia said that the convention was against its new constitution,}} adopted in 2009, which says it must ``protect native and ancestral coca'' as part of its cultural heritage and says that coca ``in its natural state ... is not a dangerous drug.'' \\ \hline
\bf Q  &\textbf{What was the purpose of the 1961 Convention on Drugs?} \\
        &A \textbf{\textcolor{red}{Regulating international policy on drugs}} \\
        &B \textbf{\textcolor{red}{Discussing whether indigenous people in Bolivia should be allowed to chew coca leaves}} \\
        &C \textit{\textcolor{blue}{Discussing the legal status of Bolivia's constitution}}  \\
        &D Negotiating extradition agreements for drug traffickers \\
\end{tabularx}
\caption{\label{tab:annotation-example} A question example with annotations for the Advanced and Elementary versions of the paragraph (note that the complete annotation contains two additional questions and the Intermediate paragraph level). The critical span is marked in bold red. The distractor span is marked in italic blue.}
\end{table*}

We implemented the STARC annotation framework in a new reading comprehension dataset, OneStopQA. The textual materials of OneStopQA are drawn from the OneStopEnglish corpus \cite{vajjala2018onestop}, which contains Guardian News Lessons articles from the English language learning portal \href{http://www.onestopenglish.com}{onestopenglish.com} by Macmillan Education. We chose articles that have non-repetitive content, and collectively represent a diverse range of topics. The texts were cleaned from errors stemming from the conversion process from the original PDFs to plain text, and manually converted from British to American English spelling. 

Each article has three versions, corresponding to three text difficulty levels: Advanced, Intermediate and Elementary. The Advanced version is the original Guardian article. The Intermediate and Elementary articles are simplified versions of the original article created by professional editors at \href{http://www.onestopenglish.com}{onestopenglish.com}. Common simplifications include text removal, sentence splitting and text rewriting. In a few cases, the edits also include changes to the presentation order of the content.
 
 OneStopQA has 30 articles, with 4 to 7 paragraphs per article, and a total of 162 paragraphs. Each paragraph has 3 to 12 sentences. Further statistics on OneStopQA and RACE articles along with readability estimates for the different text difficulty levels are presented in Table~\ref{tab:onestop-stats-table}. We note that OneStopQA paragraphs are considerably shorter than RACE articles. At the same time, even the Elementary version of OneStopQA has longer sentences and higher text difficulty level compared to the High School version of RACE. 
 We composed three reading comprehension questions for each paragraph, resulting in 486 questions, and 1,458 question-paragraph pairs when considering all three text versions. All the questions are answerable based on any of the three difficulty levels of the paragraph. Furthermore, the questions are local to the paragraph; they are answerable without any additional information from the preceding nor the following paragraphs. All the spans were annotated manually for each question in all three versions of the paragraph. Two of the questions have the same or substantially overlapping critical spans, and the third question has a distinct critical span. No restrictions were imposed on the distractor spans. Statistics for the questions, answers and spans are presented in Table~\ref{tab:answers}. Table~\ref{tab:annotation-example} presents an annotated question for two paragraph difficulty levels. Appendix \ref{sec:dataset-construction} contains details on the dataset development and piloting process.

\begin{table}[ht]
\small
\begin{center}
\begin{tabular}{c|llll}
\hline
\bf & \bf Definition & \bf Answer &\bf  Span &  \bf Span\\
\bf        &  &  \bf Length & & \bf Length\\ \hline
A & correct & 7.2 (3.5) & \multirow{2}{*}{critical} & \multirow{2}{*}{37.9 (16.5)} \\
B & incorrect & 7.6 (3.6) & &\\
C & incorrect & 8.1 (3.8) & distractor& 15.5 (11.8)\\
D & incorrect & 6.9 (3.1) & N/A & N/A\\
\hline
\end{tabular}
\end{center}
\caption{\label{tab:answers} STARC answer structure, and mean length (in words) of answers and spans in OneStopQA (standard deviation in parentheses). 50\% of the A spans comprise of more than one sentence. The mean OneStopQA question length is 11.2 words. In RACE, the mean question length is 10.0 and the mean answer length is 5.3.}
\end{table}

\begin{table*}[ht!]
\small
\begin{center}
\begin{tabular}{l|ll|c|lll|c|lll|c} 
\hline
& \multicolumn{3}{c|}{\bf RACE} & \multicolumn{4}{c}{\bf OneStopQA (no finetuning) }& \multicolumn{4}{|c}{\bf OneStopQA} \\ \cline{2-12}
        & Mid   & High & All & Ele & Int &  Adv & All & Ele & Int & Adv & All\\ \hline
Sliding Window & 41.2 & 31.0 & 33.9 & 25.6 & 26.2 & 27.5 & 26.7 & 27.7 & 27.2 & 27.3 & 28.2 \\
Stanford AR & 40.0 & 43.9 & 42.8  & 30.2 & 30.1 & 30.1 & 30.2 & 34.2 & 34.3 & 34.3  & 34.3\\ 
RoBERTa Base& 73.2 & 66.4 &  68.4  & 69.5 & 69.1 & 67.7 & 68.8 & 68.7 & 69.1 & 68.5 & 68.8  \\
RoBERTa Large & 86.6 & 81.3 & 82.9 & 85.6 & 85.0 & 86.0 & 85.6 & 86.0 & 85.4 & 86.4 & 86.0 \\ \hline
\end{tabular}
\end{center}
\caption{\label{tab:qa-baselines} QA Accuracy on RACE and OneStopQA. Random baseline on both datasets is 25.0. In ``OneStopQA (no finetuning)'' the models are trained for QA only on RACE. In ``OneStopQA'' the models are trained on RACE and further finetuned on OneStopQA.}
\end{table*}

\section{Experiments}

\begin{table*}[ht!]
\small
\begin{center}
\begin{tabular}{l|l|ll|l|lll|l|lll}
\cline{3-12}
\multicolumn{2}{c|}{} & \multicolumn{3}{c|}{\bf RACE} & \multicolumn{7}{c}{\bf OneStopQA} \\ 
\cline{3-12}
 \multicolumn{2}{c|}{} & Mid & High & All & Ele & Int & Adv & All & B & C & D\\ \hline
\parbox[t]{2mm}{\multirow{7}{*}{\rotatebox[origin=c]{90}{RoBERTa}}} & Full Information & 86.6 & 81.3 & 82.9 & 86.0 & 85.4 & 86.4 & 86.0 & 8.9 & 3.0 & 2.1 \\ \cline{2-12}
&No passage  & 46.4 & 47.3 & 47.1 & \multicolumn{3}{c|}{\cellcolor{gray!25}} & 37.2 & 19.0 & 19.9 & 23.9 \\ 
&No Q  & 61.4 & 60.6 & 60.8 & 67.7  & 68.9  & 69.9  & 68.8 & 15.5 & 13.3 & 2.4\\ 
&No Q \& No passage   & 37.8 & 40.9 & 40.0 & \multicolumn{3}{c|}{\cellcolor{gray!25}} & 34.7 & 20.0 & 20.4 & 24.9 \\ \cline{2-12}
&Only critical span & \multicolumn{3}{c|}{\cellcolor{gray!25}} & 89.3 & 86.8 & 85.8 & 87.3 & 10.4 & 0.6 & 1.7 \\
&No distractor span & \multicolumn{3}{c|}{\cellcolor{gray!25}} & 88.5 & 85.6 & 87.4 & 87.2 & 9.1 & 1.7 & 1.9 \\
&No critical span & \multicolumn{3}{c|}{\cellcolor{gray!25}} & 42.0 & 40.1 & 41.1 & 41.1 & 20.6 & 14.9 & 23.5 \\
\hline \hline
\parbox[t]{2mm}{\multirow{5}{*}{\rotatebox[origin=c]{90}{Humans}}} & Prolific QA & 85.8 & 70.3 & 74.8 & 81.7  & \multicolumn{1}{c}{-} & 79.7 & 80.7 & 10.3 & 6.8 & 2.2 \\ 
& Prolific No passage & 42.8 & 37.8 & 39.3 & \multicolumn{3}{c|}{\cellcolor{gray!25}}  & 31.9 & 21.1 & 19.5 & 27.5  \\
& Prolific \% Consensus invalid Q  & 8.0 & 22.5 & 18.3 & 2.5 &  \multicolumn{1}{c}{-} & 4.3 & 3.4
&\multicolumn{3}{c}{\cellcolor{gray!25}}\\ 
& Approximate ceiling & 94.7 & 86.4 & 88.8 & 98.5 &  \multicolumn{1}{c}{-} & 97.2 & 97.9
&\multicolumn{3}{c}{\cellcolor{gray!25}}\\ 
 & In-lab QA  & 90.7 & 82.2 & 84.7 & 96.3 & \multicolumn{1}{c}{-} & 94.4 & 95.3 & 2.3 & 1.9 & 0.5 \\ 
\end{tabular}
\end{center}
\caption{\label{tab:ablations-and-human} Ablation experiments using RoBERTa Large and Human reading comprehension experiments.}
\end{table*}

We report a series of experiments which assess human and machine reading comprehension on OneStopQA and compare it to RACE. We further showcase the ability of our annotation framework to support automated dataset quality validation and enable in-depth comparisons between human and machine reading comprehension behavior. 

\subsection{Benchmarking Machine Reading Comprehension Performance}

In this experiment, we benchmark two neural reading comprehension models, the Stanford Attentive Reader (AR) \cite{chen2016}, and RoBERTA \cite{roberta2019} a state-of-the-art model on RACE. We train the models on RACE, and evaluate their accuracy on RACE and OneStopQA. To reduce the impact of potential domain differences, we also provide an evaluation in which we further finetune the models on OneStopQA with 5-fold cross validation, where in each fold 18 articles are used for training, 6 for development and 6 for testing. Additionally, we report the performance of the commonly used sliding window baseline \cite{richardson2013mctest}. In parallel with the two neural model evaluation regimes for OneStopQA, we perform two evaluations for this baseline, one in which the window size is optimized on the RACE development set, and one in which it is optimized on OneStopQA using 5-fold cross validation.

Table~\ref{tab:qa-baselines} presents the results of this experiment. We observe that the two weaker models, Sliding Window and Stanford AR, perform better on RACE than on OneStopQA. Particularly notable is the large drop in the performance of Stanford AR from 42.8 on RACE to 34.3 on OneStopQA ($p\ll.001$, t-test). This suggests that OneStopQA is more robust to simple word-matching heuristics. The results for RoBERTa are comparable on OneStopQA and on RACE. We note that overall this is a strong outcome for OneStopQA in light of its span-based format, shorter paragraphs, and higher human ceiling performance which we discuss in Section \ref{subsec:human_reading_comprehension}. 
We further note that finetuning on OneStopQA preserves or improves performance across models by a small margin. Finally, the difficulty level of OneStopQA paragraphs has only a small and inconsistent effect on model performance.

\subsection{Ablation-based Data Quality Probing}
\label{subsec:ablations}

We introduce a new methodology for analyzing the quality of reading comprehension datasets through ablation studies. This methodology enables evaluating the robustness of OneStopQA to guessing heuristics and the validity of the relation between the answers and the span annotations. In each ablation study, we train and evaluate the performance of RoBERTa without a part of the textual input. 

The ablation studies are divided into two groups:
\begin{itemize}
    \item \textbf{Full component ablations}, applicable to any multiple choice reading comprehension dataset. In these experiments we withhold either the question, the passage or both during the training and testing of the model.
    \item \textbf{Span ablations}, which are enabled by the STARC annotations and hence apply only to OneStopQA. In the span ablation experiments we remove parts of the passage according to the span markings. These experiments enable empirical validation of the relation between answers and spans.
\end{itemize}
We report the results of these ablation studies in the RoBERTa portion of Table~\ref{tab:ablations-and-human}. 

\subsubsection*{Full component ablations}
When removing the passage, we obtain an accuracy of 37.2\% on OneStopQA, and comparable choice rates among the distractors. This is a key result which suggests that RoBERTa is not able to recover substantial information about the correct answer without the passage and provides evidence for the a-priori plausibility of all three distractor types. In contrast to this outcome, on RACE, the passage ablation experiment yields a significantly higher accuracy of 47.1 ($p\ll0.001$, t-test). The ability of RoBERTa to guess the correct answers to nearly half of the questions in RACE without requiring the passage leads to a credit assignment issue, where 22\% of RoBERTa's performance on this dataset could in principle be attributed to question and answer patterns rather than reading comprehension. 

We next exclude the question and find that OneStopQA is less robust than RACE in this regime, with an accuracy of 68.8 compared to 60.8 ($p<0.001$, t-test). This result is likely reflecting the fact that unlike in RACE, the correct answer in OneStopQA is always stated or can be directly inferred from the passage. We note that compared to the no-passage ablation, the presence of the passage eliminates D as expected. Interestingly, the relative choice rate for C is high for the no-question ablation compared to the full model, suggesting that RoBERTa is able to rule out C only in the presence of the question. This is a desirable behavior, consistent with the requirement for the C distractor to contain information from the passage which could be possibly correct, while not being a correct answer to the question. Finally, 40.0 percent of the RACE questions are guessable even when both the question and the passage are not provided, compared to 34.7 for OneStopQA ($p\ll0.001$, t-test).

\subsubsection*{Span ablations}

In the OneStopQA span ablation experiments, providing RoBERTa only with the critical span makes it focus on A and B as the only viable options, as expected. A similar C elimination outcome is obtained when the ablation is targeted at the distractor span only. 
Finally, removing the critical span, which should make the question unanswerable, results in a sharp drop in performance to an accuracy of 41.1, only 3.9\% above withholding the entire passage. Interestingly, the selection rate of C is lower compared to the full passage ablation, an outcome we intend to investigate further in the future. Overall, these results confirm the robustness of OneStopQA to guessing as well as the tight correspondence between answers and spans. We envision extending this framework in the future for automatic identification of specific items with problematic annotations which could substitute item pilots with human subjects.

\subsection{Human Reading Comprehension}
\label{subsec:human_reading_comprehension}

In these experiments we assess human reading performance and guessing behavior, and further investigate OneStopQA and RACE question quality.\footnote{The human subject data was collected under MIT IRB protocol \#1605559077 - ``Cognitive Foundations of Human Language Processing and Acquisition''. All subjects provided written consent prior to participation.} 
    \begin{itemize}
        \item \textbf{Question Answering (QA)} This experiment benchmarks human question answering performance. Participants are presented with a passage along with a question and its four answers, and are asked to select the correct answer based on the passage. After confirming their selection, participants are informed on whether they answered correctly and shown the correct answer.
        \item \textbf{Guessing (No Passage)} The goal of this experiment is to determine the extent to which humans can guess the correct answer to questions without reading the passage. Participants see only the question and its four answers and are asked to provide their best guess for the correct answer. After confirming their selection, participants are informed on whether it was correct and shown the correct answer along with the passage.
        \item \textbf{Question Validity Judging} This experiment is designed to identify questions which do not have a unique correct answer. Participants are presented with the question, answers and the passage, and are asked to indicate whether the question has (A) one correct answer, (B) more than one correct answer, or (C) no correct answer. If (A) is selected, the participant further selects the correct answer. If (B) is selected, the participant is asked to mark all the answers that they consider to be correct.
    \end{itemize}
    
We deployed all three experiments on the crowdsourcing platform Prolific (\href{https://www.prolific.co/}{prolific.co}), with a 6 trials batch for each subject. The first two trials were fixed practice items, one with a passage from OneStopQA and one from RACE. These trials were tailored for each experiment such that performing the respective task correctly is straightforward. Next, each participant performed 4 experimental trials. Two of the trials had passages from OneStopQA (one Advanced and one Elementary, taken from different articles), and two were from RACE (one Middle School and one High School). To encourage participants to perform the tasks well, in the QA and Guessing experiments participants received a monetary bonus for each correct answer. In all three experiments, participants who did not answer both practice trials correctly were excluded from the analysis.

The materials for each of the three Prolific experiments are 1296 question-passage pairs, 648 from OneStopQA and 648 from RACE. The OneStopQA items are taken from 20 OneStopQA articles, with a total of 108 paragraphs. For each paragraph we use two paragraph difficulty levels - Advanced and Elementary, combined with each of the 3 questions. The RACE materials include 108 Middle School articles and 108 High School articles from the RACE test set. We chose the articles at random among the articles that have three or more questions, and then randomly picked 3 questions for each article. In each of the three Prolific experiments we collected responses from three valid participants (i.e. participants who answered both practice trials correctly) for each question-passage pair. A single participant completed one batch in one of the three experiments, corresponding to a total of 2,916 unique participants (792 per experiment).

Even in the presence of monetary incentives and participant filtering based on practice trials, it is hard to guarantee that crowd-sourcing workers are always performing the given task attentively. We therefore further ran the QA experiment with in-lab participants.  For this experiment, we used a subset of 432 questions from the Prolific experiments' materials. 
We recruited 12 participants (6 undergraduate students and 6 post-graduate students), each completing 36 items. The items given to each participant were equally distributed between datasets and text difficulty levels, and guaranteed not to repeat the same article for RACE and the same paragraph for OneStopQA. 
The results of the human reading comprehension experiments are presented in the ``Humans'' portion of Table~\ref{tab:ablations-and-human}. Comparisons were calculated using Satterthwaite's method applied to a mixed-effects model that treats subjects and questions as crossed random effects.
All the experiments suggest clear advantages of OneStopQA as compared to RACE. In the Prolific QA experiment, participants obtain a higher overall accuracy of 80.7 on OneStopQA compared to 74.3 on RACE ($p<0.001$). We note that our QA experiment reproduces the Mechanical Turk experiment in \cite{lai2017race}, which yielded a similar human performance of 73.3 on RACE.  In the Guessing experiment, we observe that without exposure to the passage, participants were able to obtain an accuracy of 32.1 on OneStopQA as compared to 39.5 on RACE ($p\ll0.001$). For the Question Validity Judging experiment we report the percentage of questions on which all three participants have indicated that the question does not have a unique answer. This metric reveals a dramatic advantage of OneStopQA, with 3.4\% of invalid questions as compared to 18.3\% for RACE ($p\ll0.001$). We note that this result is substantially different from the percentage of invalid questions reported in Lai et al. \shortcite{lai2017race}, where the authors have estimated that only 5.5\% of the RACE questions are invalid. 

The judging experiment also enables us to devise a heuristic for approximating the ceiling performance on both datasets. To calculate it, we assign valid questions with a score of 1, and invalid questions with a score of 1 divided by the average number of answers considered correct across participants (where no correct answer is treated as 4 correct answers). The resulting performance ceiling is 88.8 for RACE and 97.9 for OneStopQA. The QA accuracy of our in-lab participants approaches this ceiling with 95.3 accuracy on OneStopQA versus 84.7 on RACE ($p<0.01$). The combination of this outcome with the results of our Question Validity experiment suggests that the human gap from perfect 100\% accuracy on RACE is due mainly to poor item quality rather than high item difficulty.

These results have important implications on current machine reading evaluations. With an accuracy of 82.9\% for RoBERTa and even higher performance for ensemble models reported on the RACE public leader board, it is likely that current machine reading models are very close to exhausting the space of meaningful performance improvements on this dataset. On the other hand, a more substantial room for improvement is still available for OneStopQA. 

\subsection{Comparing Humans and Machines}

Our final analysis uses the structured annotations of OneStopQA for detailed comparisons of human and machine reading comprehension behavior. In particular, the annotations enable comparing the error distributions of humans and machines. Interestingly, we observe that the Prolific QA error distribution is similar to that of RoBERTa, where B is the most common error, C is the second most common error and D is the least common error. This error frequency order is in line with the strength order design of the distractors. Further, similarly to RoBERTa, humans are only slightly affected by the difficulty level of the paragraph, although differently from RoBERTa, human performance is consistently worse on the advanced level compared to the elementary level. These results suggest deeper parallels between human and machine reading comprehension behavior than previously observed via overall accuracy comparisons.

Our no-passage guessing experiment on the other hand suggests interesting differences between humans and RoBERTa. First, RoBERTa, which is specifically trained on this task, has a higher guessing performance than humans on Prolific. Further, the overlap in the questions successfully guessed by humans and by RoBERTa is fairly small: the percentage of questions correctly guessed by both humans and RoBERTa is 18\% for RACE and 12\% for OneStopQA. We hypothesize that these results are due at least in part to RoBERTa picking up on statistical regularities in the question and answer training data which are difficult for humans to spot at test time. The STARC annotations enable gaining further insight into the difference in the guessing strategies of humans and machines: humans have a stronger preference for D ($p<.05$, McNemar's test). This outcome makes sense in the absence of the paragraph, as while the other answers are constrained by the specifics of the paragraph, D distractors may appeal to general world knowledge and reasoning which can be beyond the capacities of RoBERTa.


\section{Related Work}

A considerable number of reading comprehension datasets have been introduced in NLP. A large fraction of these datasets can be broadly divided into three tasks: Cloze \cite{hermann2015cnn,hill2015goldilocks,bajgar2016books}, span identification QA \cite{rajpurkar2016squad, nguyen2016ms, trischler2017newsqa, joshi2017triviaqa,naturalquestions2019} and multiple choice QA \cite{richardson2013mctest,lai2017race}.

Our approach primarily falls into the third category. The basic 4-answer format we use is identical to RACE \cite{lai2017race}, which enables training models on RACE and evaluating them on OneStopQA. Our dataset is considerably smaller than RACE, but is of appropriate size for robust evaluations and error analyses. As demonstrated in this work, OneStopQA annotations are of substantially higher quality than RACE, and enable analyses which are not possible with RACE.
MCTest \cite{richardson2013mctest} was created with a similar purpose to RACE, but has a low text difficulty level suitable for 7-year-olds. 

Span identification QA is a task in which the correct answer to the question is one or more textual spans which the reader is required to mark. This task differs from multiple choice reading comprehension in its focus on information retrieval, which limits the range of question types (e.g. forces the answers to be primarily named entities) and their difficulty level. While our approach contains span annotations, our notion of span is different from that in span identification QA: spans are not considered as answers but rather as text regions that contain the critical information for the respective answer. This difference enables a higher difficulty degree and a wider scope of question types. The combination of this approach with a multiple choice answer structure which always has a span misinterpretation distractor facilitates deeper probing of text understanding and is designed to allow for more robustness to simple pattern matching.

Prior work has explored both manual and automatic auxiliary span annotations for correct answers in multiple choice QA datasets \cite{khashabi2018-multisent,wang2019-racespans}. Our framework extends such annotations to include multiple distractor types, with B distractors providing an additional guarantee that simply identifying the critical span is not sufficient for answering the question correctly. We further demonstrate the utility of our distractor structure for automatic verification of annotation quality through ablation experiments, as well as detailed error comparisons between human and machine readers.


\section{Discussion}
We introduce a new annotation framework for reading comprehension and an accompanying high-quality dataset. We leverage the novel structure of our annotations to develop a methodology for automatic validation of annotations and to perform detailed comparisons between human and machine reading comprehension. Our experiments further demonstrate substantial quality assurance issues with RACE, which are alleviated in our new dataset. Our results demonstrate the promise of our annotation framework and dataset in supporting a wide range of reading behavior analyses, as well as the feasibility of developing automated question validation tools for reading comprehension examinations for humans as exciting directions for future work.

\section*{Acknowledgments}
We thank Beining Jenny Zhang, Katherine Xiao, Margarita Misirpashayeva and Theodor Cucu for contributions to preparation of OneStopQA materials and collection of human subject data. We also thank Sowmya Vajjala for assistance with OneStopEnglish. We gratefully acknowledge support from Elemental Cognition and from NSF grant IIS-1815529, a Google Faculty Research Award, and a Newton Brain Science Award to RPL.

\bibliography{acl2020}
\bibliographystyle{acl_natbib}

\appendix
\section{OneStopQA Construction and Piloting}
\label{sec:dataset-construction}

Questions for the OneStopQA articles were written and revised in the following manner. For each article, an annotator first composed a full draft of the questions along with span annotations. A first round of revisions for all the questions was then done by a second annotator, called ``reviewer''. Subsequently, the annotator and the reviewer resolved the issues that were raised by the reviewer.

In order to identify problematic and guessable questions, the questions were then piloted on the crowd-sourcing platform Prolific. Each participant in the Prolific pilot read a two-paragraph practice article taken from the OneStopEnglish corpus, followed by a OneStopQA article. In each single trial of the experiment, participants answered one reading comprehension question about one paragraph. Each trial consisted of three pages. On the first page, participants were presented with a question and its four possible answers and were asked to provide their best guess of the correct answer. On the following page, they read the paragraph. On the third page, the question and the answers were presented again (without the paragraph), and participants were asked to select the correct answer based on the content of the paragraph. The two practice article questions were on a lower difficulty level compared to the OneStopQA questions, and were used to identify and exclude participants who were not performing the task adequately. 

We conducted the Prolific pilot using the Elementary and Advanced versions of the articles, excluding the Intermediate level articles for cost efficiency. This resulted in six possible conditions for each trial, where each condition is a pairing of one of three possible questions with one of two possible difficulty levels for the paragraph. We consequently created 6 experimental lists with trials assigned at random to one of these conditions in a Latin square design. We collected data from 96 participants per article equally distributed between the 6 lists. This corresponds to 32 participants for each question: 16 for the Elementary version of the paragraph and 16 for the Advanced version.

The results of the Prolific pilot were used to inform a third round of revisions, which focused on questions which which fell under a set of criteria designed to facilitate the identification of guessable and problematic questions, as well as questions that catered to a specific difficulty level of the paragraph. The different criteria, along with their motivation are presented in Table~\ref{tab:prolific-revisions}. In a fourth round of revisions, the answers were edited to ensure roughly equal average lengths for the four answer types across questions. Finally, the texts, questions and answers were proofread, and the span annotations were verified.

\begin{table}[h]
\begin{center}
\resizebox{\columnwidth}{!}{\begin{tabular}{l|l|l|l}
\hline
\bf Answer & \bf Choice Rate & \bf Reading & \bf Potential Issues \\ \hline
A & $>$ 60\% & pre  & guessable\\ 
A & $<$ 50\% & post & question/answers \\
A & $>$ 95\% & post & question too easy \\
B / C / D & $>$ 30\% & post & distractor \\
Any & $>$ 30\% $|$ele - adv$|$ & post & question/answers  \\
    &                    &  &may cater to one level\\\hline
\end{tabular}}
\end{center}
\caption{\label{tab:prolific-revisions} Criteria for targeted question editing based on per question results from a crowd-sourcing pilot on Prolific.}
\end{table}

\end{document}